\documentclass[10pt,journal,compsoc]{IEEEtran}

\ifCLASSOPTIONcompsoc
  \usepackage[nocompress]{cite}
\else
  \usepackage{cite}
\fi

\usepackage{times}
\usepackage{epsfig}
\usepackage{graphicx}
\usepackage{amsmath}
\usepackage{amssymb}
\usepackage{booktabs} 
\usepackage{threeparttable}

\begin{document}

\title{A Real-time Global Inference Network for One-stage Referring Expression Comprehension}

\author{Yiyi~Zhou,
        Rongrong~Ji*,
    	Gen~Luo,
    	Xiaoshuai~Sun,
    	Jinsong~Su,
    	Xinghao~Ding,
    	Chia-Wen Lin,
    	Qi~Tian
\IEEEcompsocitemizethanks{\IEEEcompsocthanksitem Y. Zhou, R. Ji, G. Luo, X. Sun, J. Su and X. Ding are with the School of Informatics of Xiamen University, Xiamen, China.\protect
\IEEEcompsocthanksitem Corresponding Author: Rongrong Ji (E-mail: rrji@xmu.edu.cn)
\IEEEcompsocthanksitem C. Lin is with National Tsing Hua University
\IEEEcompsocthanksitem T. Qi is with University of Texas at San Antonio}
\thanks{Project: https://github.com/luogen1996/Real-time-Global-Inference-Network}}

\IEEEtitleabstractindextext{%
\begin{abstract}
Referring Expression Comprehension (REC) is an emerging research spot in computer vision, which refers to detecting the target region in an image given an text description. 
Most existing REC methods follow a multi-stage pipeline, which are computationally expensive and greatly limit the application of REC. 
In this paper, we propose a one-stage model towards real-time REC, termed Real-time Global Inference Network (RealGIN).
RealGIN addresses the diversity and complexity issues in REC with two innovative designs: the Adaptive Feature Selection (AFS) and the Global Attentive ReAsoNing unit (GARAN). 
AFS adaptively fuses features at different semantic levels to handle the varying content of expressions. 
GARAN uses the textual feature as a pivot to collect expression-related visual information from all regions, and thenselectively diffuse such information back to all regions, which provides sufficient context for modeling the complex linguistic conditions in expressions.
On five benchmark datasets, {i.e.,} RefCOCO, RefCOCO+, RefCOCOg, ReferIt and Flickr30k, the proposed RealGIN outperforms most prior works and achieves very competitive performances against the most advanced method, \emph{i.e.,} MAttNet.
Most importantly, under the same hardware, RealGIN can boost the processing speed by about 10 times over the existing methods.

\end{abstract}
}
\maketitle

\IEEEdisplaynontitleabstractindextext

%

\IEEEraisesectionheading{\section{Introduction}\label{sec:intro}}

\IEEEPARstart{R}{effering} Expression Comprehension (REC), also refers to \emph{Phase Grounding (Locating)} \cite{chen2017query-guided,plummer2017phrase,rohrbach2016grounding} or \emph{Visual Grounding} \cite{yu2018rethinking,zhang2017discriminative}, is a text-based object detection task.
It aims to locating the target region in an image based on a natural language query, \emph{e.g.,} ``\emph{man in coat and pants}''. 
Recently, REC has drawn great attention in the computer vision community~\cite{hu2017modeling,hu2016natural,krishna2018referring,nagaraja2016modeling,rohrbach2016grounding,zhang2017discriminative,zhang2018grounding}, and its advancement has been supported by a series of benchmark datasets~\cite{kazemzadeh2014referitgame:,krishna2016visual,plummer2015flickr30k} and methods~\cite{chen2017query-guided,hu2016natural,hu2017modeling,plummer2017phrase,yu2017a,yu2018mattnet:}.

\begin{figure}[t]
	\centering
	\includegraphics[width=1\linewidth]{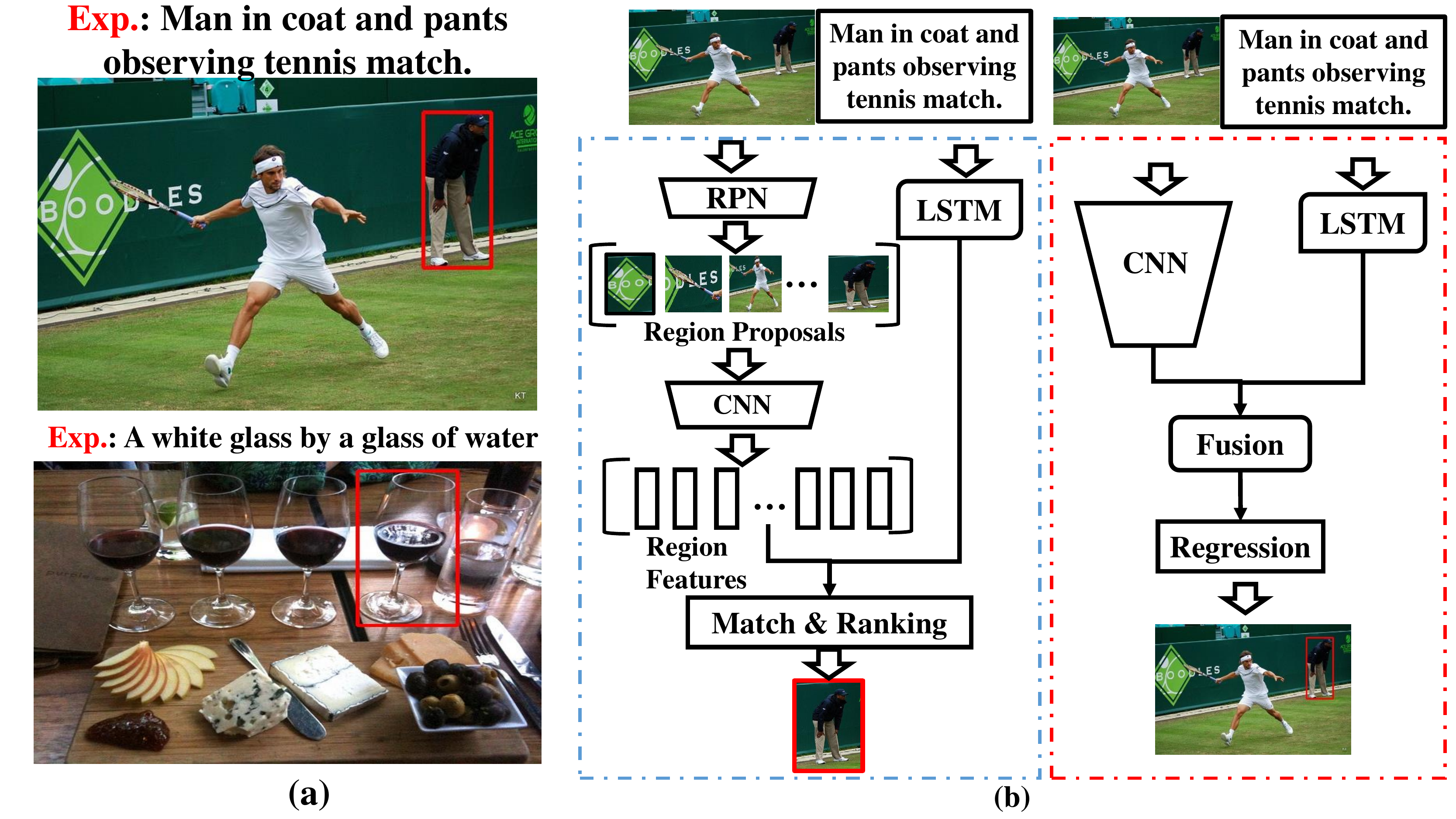}
	\caption{(a) Examples of Referring Expression Comprehension. (b) Procedures of existing methods (left) and the proposed baseline model (right).}
	\label{fig:figureoneillustration}
\end{figure}

Most existing REC methods follow a multi-stage pipeline \cite{hu2017modeling,hu2016natural,liu2017referring,luo2017comprehension-guided,mao2016generation,yu2018mattnet:,yu2017a,yu2018rethinking,zhang2017discriminative}.
As shown in Fig.\ref{fig:figureoneillustration}-(b), the core steps include: 1) generating regional proposals by using the region proposal network (RPN) \cite{ren2017faster} \footnote{Some method will directly use ground truth proposals provided by the dataset.}; 2) extracting visual\&textual features with convolutional neural networks (CNN) and language encoders such as LSTM \cite{hochreiter1997long}, respectively; 3) and then ranking the matching degree of each region-query pair to determine the target regions. 
To further enhance the efficiency of the language-vision matching, some recent models, \emph{e.g.,} CMN\cite{hu2017modeling} and MAttNet \cite{yu2018mattnet:}, also implement  additional modules for obtaining features of locations or relationships.

Although the complex multi-stage process can guarantee a relatively high performance of REC, it is also computation-ally expensive, \emph{e.g.,} about 3.0 FPS (frame per second) \cite{hu2017modeling,yu2017a,yu2018mattnet:,zhang2017discriminative}, which poses a huge obstacle to a lot of practical applications such as \emph{video surveillance} and \emph{text-to-image retrieval}. 
In addition, these multi-stage approaches rely on pre-trained object detectors, \emph{e.g.,} Faster RCNN~\cite{ren2017faster}, to obtain candidate regions, which essentially limits them to a fixed set of object categories that the detector was trained on~\cite{sadhu2019zero-shot}. 

Differing from prior work, we argue that fast and efficient REC can be achieved by a one-stage inference, formulating the task as a bounding box regression problem instead of the conventional region-query ranking problem.
We tackle the real-time requirement by enforcing the model to be trained in an end-to-end fashion and directly output the target bounding box. 
To validate this intuition, we first introduce a baseline model in Sec.\ref{sec:baseline}, which embeds a language module directly to a pre-trained object detection model, \emph{e.g.}, \emph{YOLO3}~\cite{redmon2018yolov3:}, as illustrated in Fig.\ref{fig:figureoneillustration}-(b).
Another appealing property of one-stage REC is that when the language-vision alignment is well modeled, the model can theoretically perform any fine-grained detection for unseen categories without additional training examples.

In this paper, we also identify two main issues that limit the upper bound of the one-stage REC, which are
(1) The diversity of referring expressions:  
The expression content in existing datasets~\cite{kazemzadeh2014referitgame:,mao2016generation} varies vastly, which includes various concepts like \emph{locations}, \emph{appearances}, \emph{fine-grained object categories} and \emph{actions} \emph{etc.}
In this case, the language-vision alignment requires visual features at different semantic levels, which can not be satisfied only using the last convolution layers. 
(2) The complexity of expressions: Some expressions may contain multiple conditions to indicate the referent, as shown in Fig.\ref{fig:figureoneillustration}-(a), which requires the model to perform global comparisons and inferences over the entire image.

\begin{figure*}[t]
	\centering
	\includegraphics[width=1\linewidth]{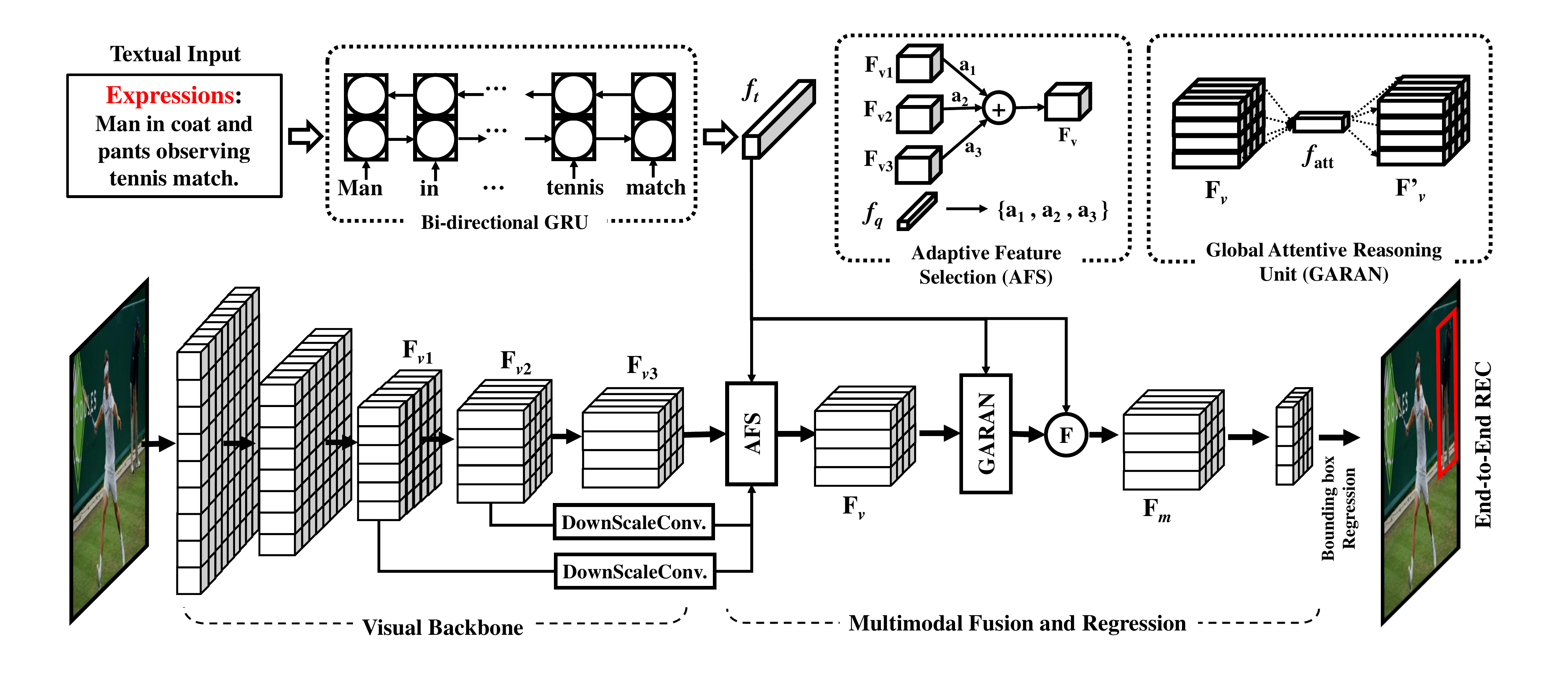}
	\caption{The framework of the proposed \emph{Real-time Global Inference Network (RealGIN). $\mathbf{F}_{v1},\mathbf{F}_{v2}$ and $\mathbf{F}_{v3}$} are feature maps of different scales, which are dynamically fused by AFS to obtain $\mathbf{F}_v$. 
		``\emph{DownScaleConv.}'' denotes the convolutional operations used for projecting $\mathbf{F}_{v1}$ and $\mathbf{F}_{v2}$ to the same dimension as $\mathbf{F}_{v3}$. 
		$f_t$ denotes the textual feature of the expression.
		Based on $f_q$, the GARAN unit is used to perform global reasoning on $\mathbf{F}_v$ to get the new feature map $\mathbf{F}_v'$, upon which the multimodal matrix $\mathbf{F}_m$ is obtained by fusing it with $f_q$.
	} 
	\label{fig:figuretwofrmework}
\end{figure*}

To this end, we propose a Real-time Global Inference Network, termed RealGIN, with two innovative designs to address the above issues. The framework is shown in Fig.\ref{fig:figuretwofrmework}.
To address the diversity of expressions and enhance the discriminability of visual features, RealGIN adopts a novel \emph{Adaptive Feature Selection} (AFS) scheme to adaptively fuses features at different semantic levels based on the expression content.
To release the complexity of expressions, we propose an innovative module called \emph{Global Attentive ReAsoNing} unit (GARAN), which uses the textual feature as a pivot to collect and diffuse context information, and performs global reasoning towards the conditions in expressions. 
Meanwhile, we also use the ground truth bounding box to achieve differentiable attentions in the GARAN unit, which subsequently boosts the model performance.

The proposed RealGIN is evaluated on three widely-used benchmarks, \emph{i.e.,} RefCoCo~\cite{kazemzadeh2014referitgame:}, RefCoCo+~\cite{kazemzadeh2014referitgame:} and RefCoCog~\cite{mao2016generation}, respectively. Very competitive performances against the most advance multi-stage SoA methods, \emph{e.g.,} MAttNet \cite{yu2018mattnet:}, are obtained.  
In some specific tasks, RealGIN even obtains distinct improvements, \emph{e.g.,} +10\% on \emph{Test A} of RefCOCO. 
Most importantly, RealGIN achieves real-time processing (26 FPS),  which is about 10 times faster than the current SoA methods. 
Conclusively, the main contributions of this paper are:

\begin{itemize}
	\item We propose a real-time and one-stage REC model to address the computation inefficiency and the generalization ability of REC, termed Real-time Global Inference network (RealGIN). 
	\item We address the issue of the expression diversity with an AFS unit, which enhances the descriptive power of RealGIN by adaptively fusing features of different semantic levels according to the textual content. 
	\item We address the issue of expression complexity with the GARAN unit, which performs a global reasoning over the entire image to handle the linguistic conditions in expressions. Notably, GARAN unit can also be generalized to most multi-modal tasks. 
\end{itemize}

\section{Related Work}

Referring expression comprehension (REC) is a task of grounding (locating) target regions in an image based on the given natural-language expression.
Existing methods regard REC as a task of selecting the best region from a set of proposals/objects based on the given expression~\cite{chen2017query-guided,plummer2017phrase,hu2016natural,hu2017modeling,yu2017a,yu2018mattnet:}. 
To this end, the region proposal network (RPN) \cite{ren2017faster} is typically adopted to generate object proposals from the given image, and CNN is used to extract the corresponding visual features.
In some settings, object proposals provided by the datasets~\cite{kazemzadeh2014referitgame:,luo2017comprehension-guided} are used.
In this work, we focus on the automatic generation of object bounding boxes. 

In terms of the methodology used, existing studies can be divided into two broad categories. 
The first is to learn a joint embedding network for two modalities and compute $P\left(r|o\right)$ of each region-query pair, where $r$ denotes the expression and $o\in O=\{o_1,...,o_i\}$ denotes the object~\cite{rohrbach2016grounding,wang2016learning,liu2017referring,chen2017query-guided,zhang2018grounding}.
Rohrbach \emph{et al.}~\cite{rohrbach2016grounding} proposed an \emph{LSTM} network to encode the expression to obtain the textual feature, which is further used as reference information to perform \emph{visual attention}~\cite{xu2015show} on region features. The obtained attention scores are further used to indicate the relevance between expression and region. Wang \emph{et al.}~\cite{wang2016learning} proposed a deep {Canonical Correlation Analysis} (CCA)~\cite{hardoon2004canonical} network to learn the joint embedding space to rank each region-query pair. 
Based on the model proposed in \cite{rohrbach2016grounding}, Liu \emph{et al.}~\cite{liu2017referring} added two attribute classifiers to enhance the descriptive ability of visual features, and the predicted attributes are further used as additional features for ranking each region-expression pair.

The other is to model the task as $P\left(o|r\right)$, which looks for the object $o$ given the expression $r$ that maximizes the probability~\cite{mao2016generation,yu2016modeling,nagaraja2016modeling,hu2016natural,luo2017comprehension-guided}. 
This type of methods typically take visual region features account into the language modeling and use the last hidden state of language network, \emph{e.g.,} LSTM, to predict the relevance of each region-query pair. 
There are also some works that combine the above two strategies and regards the comprehension and generation of referring expressions as a joint task~\cite{yu2017a,mao2016generation,luo2017comprehension-guided}. 
Our model is different from the above methods by casting REC as a multi-modal bounding box regression task.

A challenging problem in REC is the modeling of context information based on the linguistic conditions in the expression, which is the key to distinguish the referent from other objects \cite{zhang2017discriminative,zhang2018grounding,hu2017modeling,yu2018mattnet:}. 
The simplest method is to use the holistic image information as additional features for ranking region-expression pairs~\cite{zhang2017discriminative}. 
Some methods compare the visual features from neighbor regions \cite{yu2016modeling,yu2017a,nagaraja2016modeling}, and use {multiple instance learning (MIL)}~\cite{dietterich1997solving} to optimize the model. In recent work, Zhang \emph{et al.}~\cite{zhang2018grounding} uses a variational Bayesian method to model the referent and context region pairs, Hu \emph{et al.}~\cite{hu2017modeling} and Yu \emph{et al.}~\cite{yu2018mattnet:} proposed modularized components to obtain features of locations and relationships in addition to the original visual region ones. However, these region-wise comparisons subsequently exacerbate the computation cost of REC.
Our work adopts a global reasoning unit to model such context information, which can be optimized by an end-to-end manner.

\section{Baseline Model}\label{sec:baseline}
We first present a baseline model to preliminarily validate the feasibility of one-stage REC. 
Unlike prior works of selecting the best matching proposal from candidates~\cite{mao2016generation,hu2016natural,hu2017modeling,yu2018mattnet:}, our baseline model predicts the bounding box directly based on the expression. 

As illustrated in Fig.\ref{fig:figureoneillustration}, the baseline model consists of two main modules: the visual backbone and the language encoder. 
The visual backbone can be any CNN network, and we use the convolutional feature maps before the last pooling layer as the visual representations, denoted by $\mathbf{F}_v\in\mathbb{R}^{s\times s\times m}$, where $s$ is the scale and $m$ is the channel number. 
The language encoder is a bi-GRU network, where the last hidden states of the forward and backward GRUs are added as the textual feature, denoted by $f_t\in\mathbb{R}^n$.

Afterwards, we project these two types of features into the same semantic space and then fuse $f_t$ with each regional feature $f_v^i\in \mathbb{R}^m$ in $\mathbf{F}_v$ to obtain a multi-modal matrix $\mathbf{F}_m\in \mathbb{R}^{s\times s\times d}$, formulated as:
\begin{equation}\label{eq:1}
f_m^i=\sigma\left(f_v^i\mathbf{W}_v\right)\odot\sigma\left(f_q\mathbf{W}_t\right).
\end{equation}
where $\mathbf{W}_v\in \mathbb{R}^{m\times d}$ and $\mathbf{W}_t\in\mathbb{R}^{n\times d} $ are the weight matrices, and $\sigma\left(\cdot\right)$ denotes any activation function. $f_m^i$ is the feature vector of $\mathbf{F}_m$. 
Based on $\mathbf{F}_m$, we add a linear convolution layer to predict the target bounding boxes. 

In terms of the bounding box prediction, we refer to the setting of Yolo3~\cite{redmon2018yolov3:}, an recent one-stage detection model. 
Specifically, the baseline model will predict $N$ bounding boxes on each anchor (cell) of the feature map. 
Each bounding box has the width and height priors, denoted by $\left(p_w,p_h\right)$, and we predict its 4 coordinates, denoted by $\left(t_x,t_y,t_w,t_h\right)$. 
Given the offset of the anchor from the top left corner of the image, denoted by $\left(c_x,c_y\right)$, the predicted bounding box $b$ corresponds to:
\begin{equation}
\begin{split}
b_x&=\sigma\left(t_x\right)+c_x,\\
b_y&=\sigma\left(t_y\right)+c_y,\\
b_w&=p_we^{t_w},\quad
b_h=p_he^{t_h}.
\end{split}
\end{equation}
Here, $\sigma\left(\cdot\right)$ denotes the Sigmoid function.
In addition to the 4 coordinates, we predict a confidence score $p$ for the identification of the target object in the predicted box, which is formulated as:
\begin{equation}
IOU\left(b,TargetObject\right)=\sigma\left(t_c\right),
\end{equation}
where $t_c$ is the prediction logit.
So, the regression layer will output a matrix with a dimension of $s\times s\times \left(N\times 5\right)$.

The loss function of the baseline model is then formulated as:
\begin{equation}
\mathcal{L}\left(t_i,p_i\right)=\sum_{i}^{s\times s\times N} p_i^*\mathcal{L}_{box}\left(t_i^*,t_i\right)+\mathcal{L}_{conf}\left(p_i^*,p_i\right),
\end{equation}
where $\mathcal{L}_{box}$ is the regression loss of bounding box, $\mathcal{L}_{conf}$ is the binary prediction loss of the confidence score, and $p_i^*$ is a binarized ground truth with an IOU threshold of 0.5~\cite{redmon2018yolov3:}.  
In $\mathcal{L}_{box}$, we use the {binary cross-entropy} to measure the regression loss of $t_x$ and $t_y$.  For $t_w$ and $t_h$, we adopt the {smooth $l\_1$}-loss\cite{ren2017faster} which is found to be more stable than the $l\_2-$loss used in \cite{redmon2018yolov3:} in our experiments. 

During the test, the model will output the bounding box with the highest confidence\footnote{It can be extended to the multi-object detection by setting a threshold of the confidence score.}.
Note that, the settings of the above bounding box prediction and loss function can be modified according to any other one-stage detection model. 

\section{Real-time Global Inference Network}
Based on the baseline model, we further propose a real-time global inference network, denoted as \emph{RealGIN}, to address the two main issues of the one-stage referring expression comprehension, \emph{i.e.,} the diversity and the complexity of expressions, as discussed in Sec.\ref{sec:intro}.
RealGIN has the same visual backbone, language encoder, prediction layer and loss function as the baseline model, and its main differences are two innovative designs embedded into the end-to-end modeling, \emph{i.e.,} \emph{adaptive feature selection} scheme and \emph{global attentive reasoning}. 

\subsection{Adaptive Feature Selection}
The main target of Adaptive Feature Selection (AFS) is to adaptively fuse visual feature maps at different semantic levels to enhance the model discriminability, thereby handling the varying content in expressions. The illustration of AFS is shown in Fig.\ref{fig:figuretwofrmework}.

Specifically, given $k$ visual feature maps from different scales, we first use convolution layers to project them to the same resolution and  depth by setting the kernel size, stride and channel number. After that, we perform a weighted sum on these maps, and the final visual feature maps $\mathbf{F}_v\in \mathbb{R}^{s\times s\times m}$ can be formulated as:
\begin{equation}
\mathbf{F}_v=\sum_{i}^{k}\beta_i\mathbf{F}_{vi}.
\end{equation}
Here, $\mathbf{\beta}$ are the weights predicted by the textual feature $f_q$.

With this setting, we are able to fuse visual information at different semantic levels to represent the varying expressions, which is found to be effective in modeling expressions about object appearances. 

\subsection{Global Attentive Reasoning}\label{sec:garan}
\begin{figure}
	\centering
	\includegraphics[width=1\linewidth]{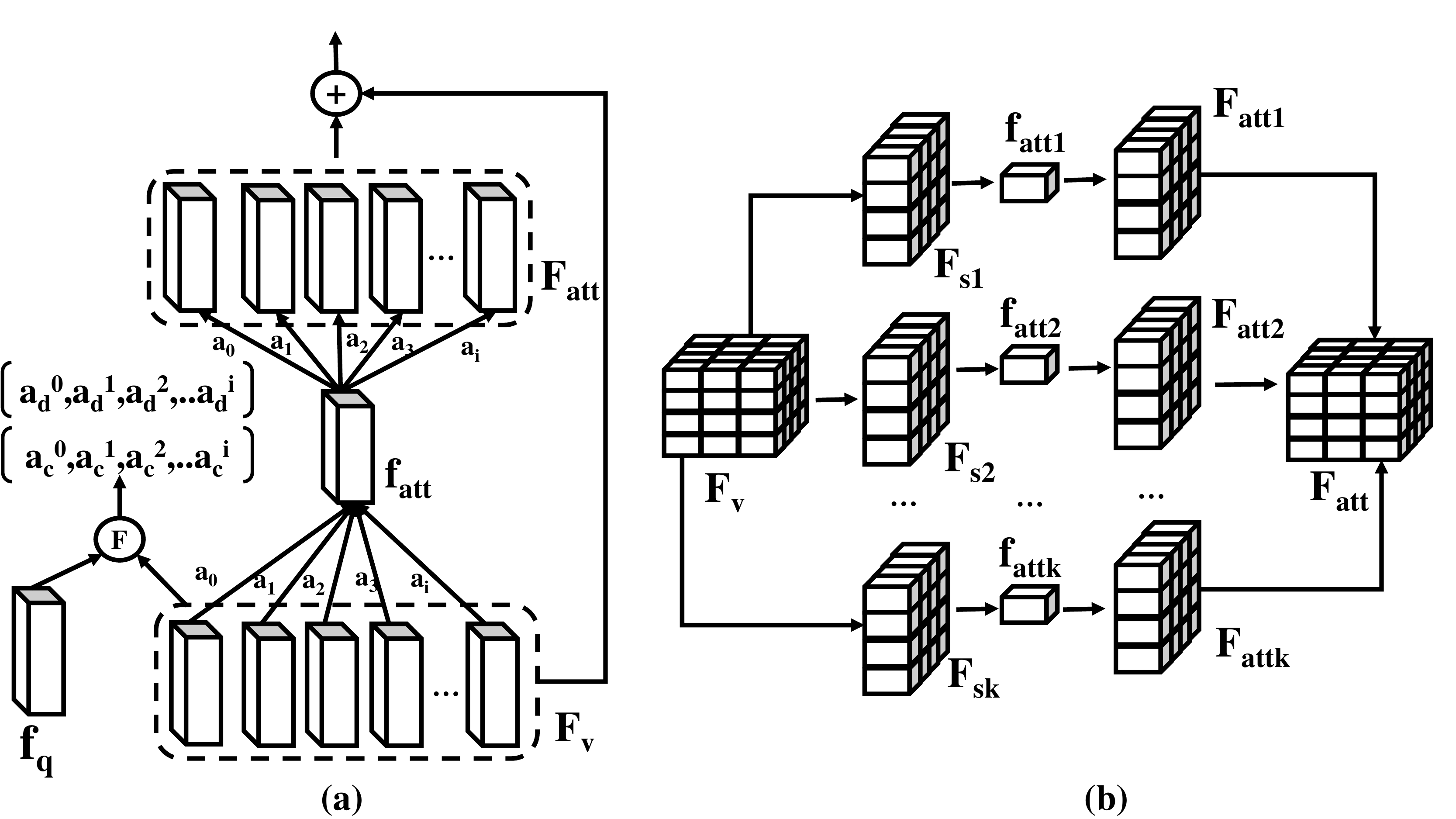}
	\caption{(a) The structure of the proposed GARAN unit. (b) The illustration of GARAN unit with multi-head attentions.}
	\label{fig:figurethreegaran}
	
\end{figure}

The key to address the complexity of referring expressions is to model context information well based on the linguistic conditions involved. 
To this end, a one-stage REC model should be able to perceive expression-related information over the entire image, and complement the intrinsic deficiency of convolutional feature maps, \emph{i.e.,} the limited receptive field of anchors in a feature map~\cite{lin2017feature}.
We achieve the above goal by proposing a Global Attentive ReAsoNing unit (GARAN). 

GARAN is an attention-based component that greatly differs from existing attention models~\cite{xu2015show,yang2016stacked,teney2018tips}, which typically use the attention feature as the visual feature \cite{xu2015show,teney2018tips} or to refine the textual feature \cite{yang2016stacked}.
In contrary, GARAN uses the textual feature as a pivot to collect expression-related information over the whole image, and then selectively diffuse this information to all anchors, as illustrated in Fig.\ref{fig:figurethreegaran}-(a).

Concretely, given the textual feature $f_t$ and the visual feature map $\mathbf{F}_v$, we fuse two types of features to predict the two attention maps $\mathbf{A}_{col}$ and $\mathbf{A}_{dif} \in \mathbb{R}^{s\times s}$ over anchors.
Here, $\mathbf{A}_{col}$ is used to obtain the attention feature, while $\mathbf{A}_{dif}$ is used in the information diffusion.
Each value $a_{c}^i$ in $\mathbf{A}_{col}$ is calculated by:
\begin{equation}\label{eq:6}
\begin{split}
a_c^i&=\frac{\exp \left(e_c^i\right)}{\sum_{j}^{s\times s} \exp\left(e_c^j\right)},\\
where\quad e_c^i&=\sigma\left(f_v^i\mathbf{W}_{va}\right)^T\cdot\sigma\left(f_t\mathbf{W}_{ta}\right).
\end{split}
\end{equation}
Here, $\mathbf{W}_{va}$, $\mathbf{W}_{ta}$ are the projection weight matrices, respectively. $e_i$ is the scalar product of two projected vectors, indicating the relevance between features of two modalities. 
Values in $\mathbf{A}_{dif}$ are obtained in the same way as $\mathbf{A}_{col}$, and the only difference is that we use {Sigmoid} function to obtain the attention weights. 
Then, we perform the weighted sum of $\mathbf{F}_v$ to obtain the attention feature $f_{att}$, which can be formulated as:
\begin{equation}
f_{att}=\sum_{i}^{s\times s}\alpha_c^i f_v^i.
\end{equation}
Afterwards, based on $\mathbf{A}_{dif}$, we convert $f_{att}$ to a feature matrix with the same dimension of $\mathbf{F}_v$, denoted by $\mathbf{F}_{att}$. The feature vector $f_a^i$ of $\mathbf{F}_{att}$ is the diluted attention feature obtained by $f_a^i=\alpha_d^if_{att}$.
Lastly, we combine $\mathbf{F}_{att}$ with $\mathbf{F}_v$ using element-wise addition and obtain the new feature maps $\mathbf{F}_v'$ with another convolution layer. 

\textbf{GARAN with multiple attentions.}
Towards a better context modelling, we draw on the multi-attention design from~\cite{vaswani2017attention} to enhance the attention ability of GARAN, as shown in Fig.\ref{fig:figurethreegaran}-(b). 

Specifically, we decompose $\mathbf{F}_v$ into $k$ splits from the last dimension, and obtains a set of feature maps, denoted by $\mathbf{F}_{s1}, \mathbf{F}_{s2}, ... \mathbf{F}_{si}\in \mathbb{R}^{s\times s\times \left(m/k\right)}$. Afterwards, we perform the above attention process on each feature map and then recombine all generated attention matrices to obtain a complete one, denoted by $\mathbf{F}_{att}\in \mathbb{R}^{s\times s\times m}$.

With such a design, RealGin will perform attentions on  a set of truncated visual features processed by different convolution filters, which can perceive more accurate and sufficient context information for bounding box regression.

\begin{figure}[t]
	\centering
	\includegraphics[width=1\linewidth]{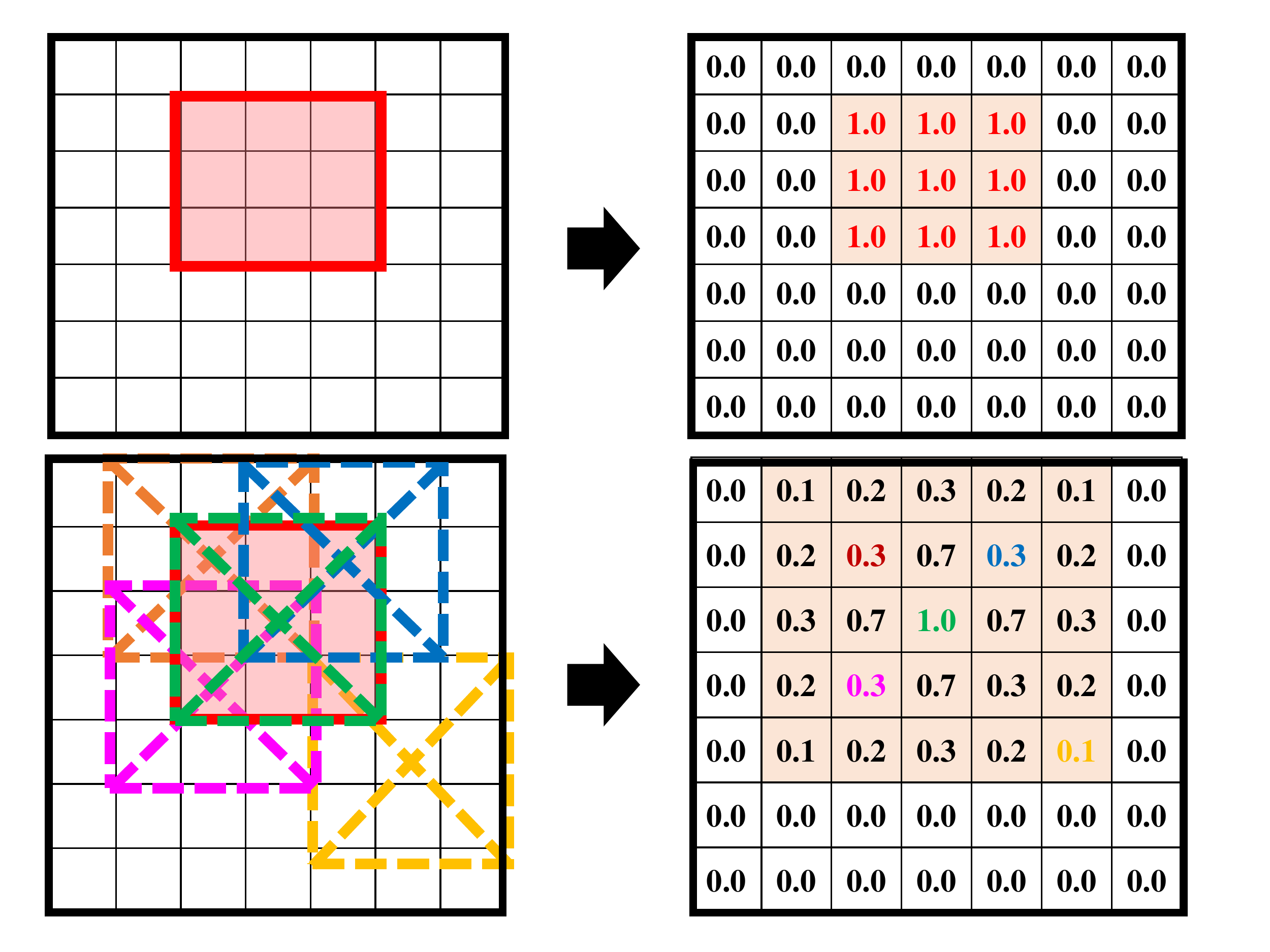}
	\caption{Two methods for converting the ground truth bounding box to the attention label.
		The first one is to directly calculate the overlapping degrees, \emph{e.g.,} IoU, between cells and the target.
		The second one used in this paper is to place the copied box on each anchor (cell) and then  calculate the overlapping degrees between the copied one and the target one.
	}
	\label{fig:att_calculate}
\end{figure}

\textbf{Differential Attentions.}
Due to the lack of corresponding annotations, previous attention-based models cannot directly optimize the attention layer~\cite{xu2015show,yang2016stacked}.
In REC, we can use the ground truth bounding box as the supervision information to achieve differential attentions. 

A natural solution for converting the bounding box to the label information is to calculate the overlapping degree, \emph{e.g.,} IoU, between each grid cell in the feature map and the given box, as shown in the top example of Fig.\ref{fig:att_calculate}. 
However, such a method will lead to a result that many anchors have the same attention score, which can not accurately reflects the importances of visual regions towards the final task. 

To this end, we propose an innovative method to calculate the ground truth scores of attentions. 
As shown in the bottom example of Fig.\ref{fig:att_calculate}, we place the bounding box with the same shape of the target one on each anchor.
Then, the attention score of an anchor is decided by the IoU value between its placed box and the target one. 
With this setting, there is at most one anchor with a score of $1.0$. 
Besides, for those anchors that are distant from the center point of ground truth box, their attention scores will decrease according to the distances, which is closer to the behavior of human cognition~\cite{xu2015show} and more suitable for our task. 

Given the ground truth score of an anchor $p_a^i$, we compute the \emph{sigmoid binary cross entropy} with the attention logit $e_i$ defined in Eq.\ref{eq:6}. Then the attention loss is formulated as:
\begin{equation}
\mathcal{L}\left(p_a^i,e_i\right)=\sum_{i}^{s\times s}\sum_{j}^{k}\mathcal{L}_{binary}\left(p_a^i,e_{ij}\right),
\end{equation}
where $k$ is the number of attention heads. 

Lastly, the overall loss function of RealGIN can be denoted as:
\begin{equation}
\mathcal{L}=\mathcal{L}\left(t_i,p_i\right)+\lambda\mathcal{L}\left(p_a^i,e_{ij}\right),
\end{equation}
where $\lambda$ is the hyper parameter.

\section{Experiments}
We further evaluate the proposed baseline model and RealGIN on five benchmark datasets, \emph{i.e.,} RecCOCO~\cite{kazemzadeh2014referitgame:}, RefCOCO+~\cite{kazemzadeh2014referitgame:}, RefCOCOg~\cite{mao2016generation}, ReferIt~\cite{kazemzadeh2014referitgame:} and Flicker 30k entities~\cite{plummer2015flickr30k},  and compare them to a set of SoA methods. 
Besides, we also validate the generalization ability of RealGIN on the Zero-Shot Grounding (ZSG) dataset~\cite{sadhu2019zero-shot}.

\subsection{Datasets and Evaluation Metric}
\textbf{RefCOCO} (UNC RefExp)~\cite{kazemzadeh2014referitgame:} has 142,210 referring expressions for 50,000 bounding boxes in 19,994 images from MS-COCO~\cite{lin2014microsoft}, which is split into \emph{train}, \emph{validation}, \emph{Test A} and \emph{Test B} with a number of 120,624, 10,834, 5,657 and 5,095 samples, respectively. 
The expressions are collected using an interactive game interface~\cite{kazemzadeh2014referitgame:}, and they are typically short sentences with a average length of 3.5 words.
The categories of bounding boxes in Test A are people while the ones in Test B are objects.

\textbf{RefCOCO+}~\cite{kazemzadeh2014referitgame:} has 141,564 expressions for 49,856 boxes in 19,992 images from MS-COCO. It is also divided into splits of \emph{train} (120,191), \emph{val} (10,758), Test A (5726) and Test B (4,889). 
Compared to RefCOCO, its expressions include more appearances (attributes) than absolute locations~\cite{kazemzadeh2014referitgame:} to describe the target box. 
Similar to RefCOCO, expressions of Test A in RefCOCO+ are about people while the ones in Test B are objects. 

\textbf{RefCOCOg} (Google RefExp) \cite{mao2016generation} has 95,010 expressions for 49,822 boxes in 25,799 images from MS-COCO. The split is 85,474 and 9,536 samples for training and validation. Since the \emph{test} set is not released, we use the \emph{UNC} partitions~\cite{yu2016modeling,yu2018mattnet:} of the \emph{val} split  for validation and testing. 
Compared to the above two datasets, expressions in RefCOCOg are collected in a non-interactive way, and the lengths are longer (8.4 words on average), of which content
includes both appearances and locations of the referent.

\textbf{ReferIt}~\cite{kazemzadeh2014referitgame:} contains 130,525 expressions referring to 96,654 distinct objects in 19,894 images of real world scenes. All the expressions are collected by the two player game. These expressions cover most objects present in the IAPR corpus.

\textbf{Flickr 30k Entities}~\cite{plummer2015flickr30k} is built based on the Flickr30k dataset, which augments about 158k captions from Flickr30k with 244k coreference chains, linking mentions of the same entities across different captions for the same image, and associating them with 276k bounding boxes.

\textbf{Metric.}
Following the setting in prior works~\cite{kazemzadeh2014referitgame:,yu2016modeling,yu2018mattnet:}, we use the precision as the evaluation metric. Specifically, when the overlapping degree, \emph{i.e.,} IoU, between the predicted bounding box and the ground truth one is larger than 0.5, then the prediction is correct. 
\begin{table}
	\caption{Ablation studies of the baseline model and RealGIN on the \emph{val} set of three datasets, where $k$ denotes the number of attentions and $\xi$ denotes pretraining on ImageNet. *is the final setting of RealGIN. }
	\centering
	\label{tab:ab_study}
	\footnotesize
	\begin{tabular}{lccc}
		\toprule
		\textbf{Setting}						&\scriptsize\textbf{RefCOCO}&\scriptsize\textbf{RefCOCO+} &\scriptsize\textbf{RefCOCOg}\\
		\hline
		Visual Only								&29.0			 & 34.8				& 39.9\\
		$^\xi$ Baseline without pretrain		&52.2			 & 36.2				& 24.0\\
		Baseline								&70.8			 & 50.7				& 59.4 \\
		\hline
		Baseline								&70.8			 & 50.7				& 59.4 \\
		Baseline+AFS							&73.2			 & 56.0				& 59.4 \\
		Baseline+GARAN \tiny{(k=1)}				&71.2			 & 51.3				& 60.1\\
		Baseline+GARAN \tiny{(k=2)}				&72.0			 & 53.0				& 60.8 \\
		Baseline+GARAN \tiny{(k=2)}+\scriptsize att\_loss	&72.9			 & 54.1				& 62.8\\
		Baseline+\scriptsize{AFS+GARAN+att\_loss*}			&77.3			 & 62.8				& 62.8 \\
		\bottomrule
	\end{tabular}
	
\end{table}
\begin{table}
	\caption{Comparisons between other context modeling scheme and the proposed GARAN on RefCOCO.  }
	\centering
	\label{tab:context}
	\footnotesize
	\begin{tabular}{lccccc}
		\toprule
		\textbf{Methods}						&&&\scriptsize\textbf{Val}&\scriptsize\textbf{Test A} &\scriptsize\textbf{Test B}\\
		\hline
		Baseline								&&&70.8			 & 72.7				& 68.4\\
		Baseline+Att							&&&71.0			 & 72.8				& 68.7 \\
		Baseline+Att+Att\_loss					&&&71.2			 & 73.4				& 69.3 \\
		Baseline+Transformer\scriptsize{(k=1)}		&&&71.2			 & 73.2				& 69.1 \\
		Baseline+Transformer\scriptsize{(k=2)}		&&&71.8			 & 73.6				& 69.5 \\
		\hline
		Baseline+GARAN\scriptsize{(k=1)}		&&&71.5			 & 73.5			& 69.6\\
		Baseline+GARAN \scriptsize(k=2)			&&&72.0			 & 73.7				& 69.7\\
		Baseline+GARAN \scriptsize(k=2)+\footnotesize{ Att\_loss}	 &&&72.9	& 74.8				& 70.0\\
		\bottomrule
	\end{tabular}
	
\end{table}
\subsection{Experimental Settings}
\begin{table*}[t]
	
	\caption{ Comparisons with the state-of-the-arts on RefCOCO, RefCOCO+ and RefCOCOg under the setting of automatic bounding box generation. }\centering
	\label{tab:sota_results}
	\centering
	\footnotesize
	\begin{tabular}{llccccccccc|c}
		\toprule
		&					&\multicolumn{3}{c}{\textbf{RefCOCO}}&\multicolumn{3}{c}{\textbf{RefCOCO+}}&\multicolumn{3}{c}{\textbf{RefCOCOg}} & \\ \hline
		\textbf{Method}& \textbf{Backbone} & \textbf{Val} &\textbf{Test A} &\textbf{Test B} & \textbf{Val} &\textbf{Test A} &\textbf{Test B} & \textbf{Val$^\Psi$} & \textbf{Val} &\textbf{Test} &\textbf{Speed (FPS)*
		}\\ \hline
		MMI\cite{mao2016generation}			& vgg16 & -		& 64.90		& 54.51				& -		&54.03		&42.81					&45.85	&-				&-&-\\
		NegBag\cite{nagaraja2016modeling} 	& vgg16 & -		& 58.60		& 56.40				& -		&-		&-					&39.50			&- 		&-&-\\
		CRE\cite{luo2017comprehension-guided}& frcnn-vgg16 & -		& 68.11		& 54.65				& -		&56.61		&43.74					&47.60	&-		&-&-\\
		CRE(w2v)\cite{luo2017comprehension-guided}& frcnn-vgg16 & -		& 67.94		& 55.18				& -		&57.05		&43.33					&49.07	&-		&-&-\\
		visdif\cite{yu2016modeling}			& vgg16 & -	& 62.50		& 50.80				& -		&50.10		&37.48					&41.85		&-	&-&-\\
		visdif+MMI\cite{yu2016modeling}		& vgg16 & -	& 67.64		& 55.16				& -		&55.81		&43.43					&46.85		&-	&-&-\\
		attr+MMI+visdif\cite{liu2017referring} & vgg16 & -		& 72.08		& 57.29				& -		&57.97		&46.20					&52.35	&-	&-&-\\
		CMN\cite{hu2017modeling} 			& frcnn-vgg16 & -		& 71.03		& 65.77				& -		&54.32		&47.76					&57.47	&-	&-&-\\
		Speaker\cite{yu2017a} 				& frcnn-vgg16 & 69.48		& 73.71		& 64.96				& 55.71		&60.74		&48.80				&59.51	&60.21			&59.63&-\\
		Listener\cite{yu2017a} 				& frcnn-vgg16 & 68.95		& 73.10		& 64.85				& 54.89		&60.04		&49.56				&57.72	&59.33			&59.21&-\\
		VariContext\cite{zhang2018grounding}& frcnn-vgg16 & -			& 73.33		& 67.44				& -			&58.40		&53.18					&62.30	&-			&-&-\\
		ParralAtt$^\xi$\cite{zhuang2018parallel}	& frcnn-res101 & -			& 75.31		& 65.52				& -			&61.34		&50.86					&58.03			&-&-&-\\
		DPPN\cite{yu2018rethinking}$^\xi$	& frcnn-vgg16	& 73.40			& 76.90		& 67.50			& 54.89		&60.04		&49.56					&-	&-		&-&3.0\\
		DPPN\cite{yu2018rethinking}$^\xi$	& frcnn-res101	& 76.08			& 80.10		& 72.40			& 64.80		&70.50		&54.10					&-	&-		&-&2.9\\
		MattNet\cite{yu2018mattnet:}		& frcnn-res101 & 76.40		& 80.43		& 69.28				& 64.93		&70.26		&56.00					&-	&\textbf{66.67}			&67.01&2.9\\
		MattNet-mrcn\cite{yu2018mattnet:}		& mrcnn-res101 & 76.65		& \textbf{81.14}		& {69.99	}		& \textbf{65.33}		&\textbf{71.62}		&56.02					&-	&66.58			&\textbf{67.27}& 2.8\\
		\hline
		Baseline (ours)						& DarkNet & 70.53		& 72.67			& 68.42		& 50.73		&52.86		&45.96						&-	&59.38			&58.89&\textbf{28.6}\\
		Baseline (ours)	$^\xi$					& DarkNet & 74.00		& 74.10			& 73.31		& 52.80		&54.50		&49.20						&-	&62.60			&63.87&\textbf{28.6}\\
		RealGIN (ours)						& DarkNet & 77.25		& 78.7			& 72.10			& 62.78		&67.17		&54.21					&-	&62.75			&62.33 &{26.3}\\
		RealGIN (ours)	$^\xi$					& DarkNet & \textbf{80.38}		& 81.08			& \textbf{77.25}			& 62.90		&{65.50}		&\textbf{57.40}						&-	&65.52			&65.57&{26.3}\\
		\bottomrule
		
	\end{tabular}
	\begin{tablenotes}
		\footnotesize
		\item $\Psi$ The Google RefCOCOg val split.
		\item $*$ We test SoA methods that release source codes on the same hardware, \emph{i.e.,} GTX1080ti. Different DL libraries used will slightly affect the speed.
		\item $\xi$ backbone pretrained with the complete MS-COCO set. 
	\end{tablenotes}
	
\end{table*}

\begin{table}
	\caption{ Comparisons with the state-of-the-arts on Refrit and Filckr30k under the setting of automatic bounding box generation. }
	\centering
	\label{tab:refit}
	\footnotesize
	\begin{tabular}{lcccccc}
		\toprule
		&&   &\multicolumn{2}{c}{\textbf{Flickr30k}}&\multicolumn{2}{c}{\textbf{ReferIt}}\\\hline
		\textbf{Methods} && \textbf{Backbone} &\textbf{Val}&\textbf{Test}&\textbf{Val}&\textbf{Test}\\
		SCRC~\cite{hu2016natural} && VGG & - &27.80 & - &17.90\\ 
		GroundeR~\cite{rohrbach2016grounding} && VGG &- &48.38 &- &28.50\\ 
		MCB~\cite{fukui2016multimodal} && VGG &- &48.70 &- &28.90\\ 
		DAP~\cite{li2017deep} && VGG &- &- &- &40.00\\  
		QRC~\cite{chen2017query-guided} && VGG & - &65.14 &- &44.10\\ 
		CITE~\cite{plummer2017conditional} && VGG &- &61.89 &- &34.13\\ 
		ZSGNet~\cite{sadhu2019zero-shot} && ResNet50 &-&63.39 &- &58.63\\ 
		ZSGNet* 						&& ResNet50 &63.15&63.43&65.99&62.73\\\hline
		RealGIN (Ours)$^\xi$ && ResNet50 & \textbf{64.53} & \textbf{65.98} & \textbf{68.29} &\textbf{65.37}\\
		\bottomrule
	\end{tabular}
	\begin{tablenotes}
		\footnotesize
		\item $*$ Reimplemented Results
		\item $\xi$ A multi-scale version of RealGIN using ZSGNet as the base model. 
	\end{tablenotes}
	
\end{table}

\begin{figure*}
	\centering
	\includegraphics[width=0.98\linewidth]{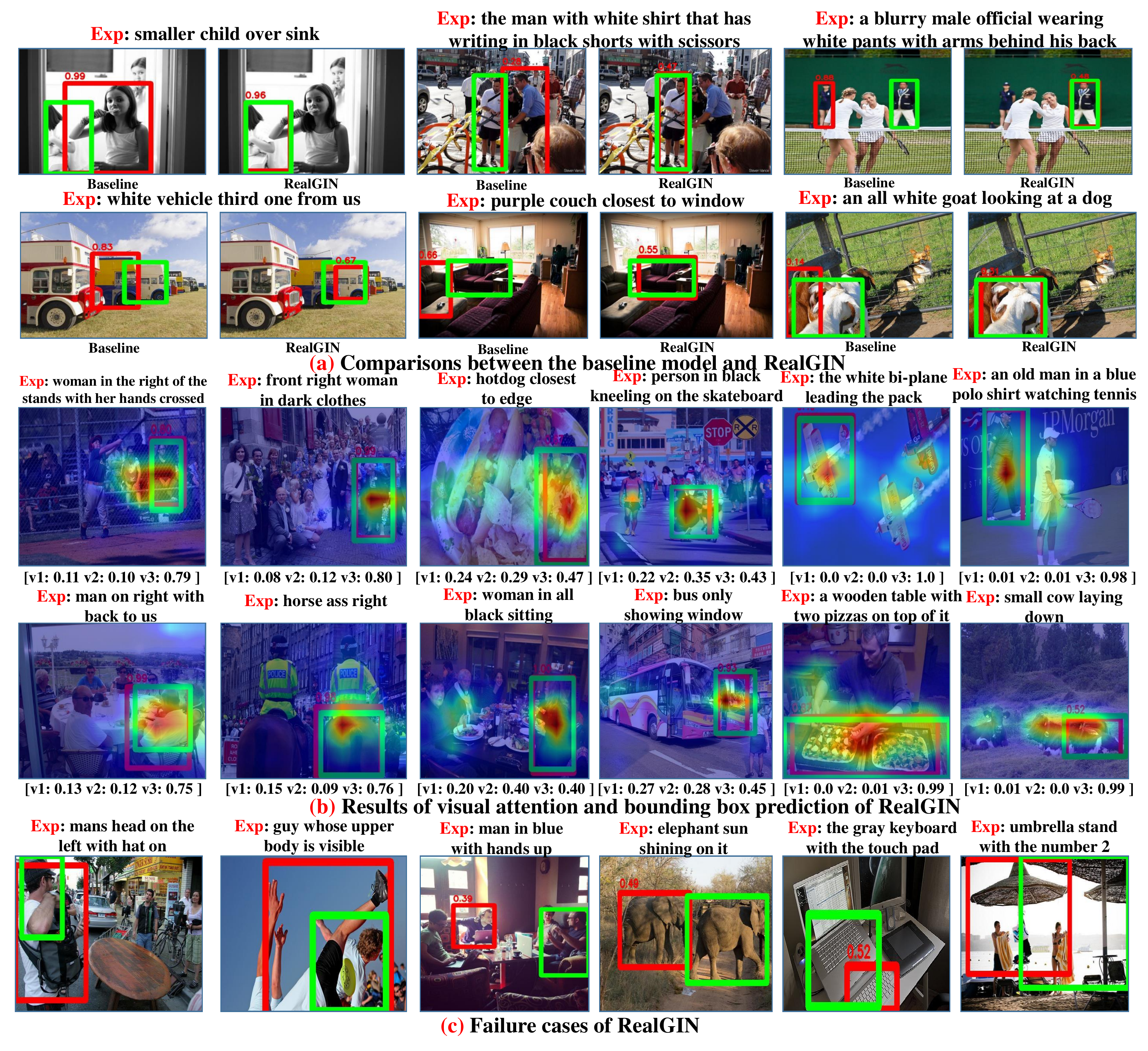}
	
	\caption{Visualizations of the baseline model and RealGIN. The green boxes are the ground truths while the red ones are the predictions.
		Values of $v$ under images are AFS scores of feature maps of three scales.
	}
	\label{fig:figurefourvis}
	
\end{figure*}
For the proposed baseline model and RealGIN, we use the Darknet from Yolo3~\cite{redmon2018yolov3:} as the visual backbone on RefCOCO, RefCOCO+ and RefCOCOg datasets.
Following the setting in ~\cite{yu2018mattnet:}, the backbone is pretrained on the MS-COCO dataset while removing the images appeared in the validation and test sets of RefCOCO, RefCOCO+ and RefCOCOg. 
Considering that some previous methods~\cite{yu2018rethinking,zhuang2018parallel} use backbones pretrained on the complete MS-COCO dataset, we also report the experiment results using the pre-trained Darknet by Yolo3~\cite{redmon2018yolov3:} for fair comparisons.
For the baseline model, we use the last convolution feature map of DarkNet as the visual features with a shape of $13\times 13 \times 1,024$. 
For RealGIN, the outputs of \emph{Layer26} and \emph{Layer43} of DarkNet are used addition-ally with a dimension of $52\times 52\times 256$ and $26\times 26\times 512$, respectively.
The kernal size and channel number of convolution layers for projecting Layer92 and Layer152 are set to be $3\times 3$ and 1,024, and the stride is $4$ and $2$, respectively.
Words in the expressions are initialized with GLOVE embeddings~\cite{pennington2014glove}.
The dimension of the forward GRU and the backbward GRU is set to be 1,024. 
In terms of multi-modal fusion, the dimension $d$ in Eq.\ref{eq:1} is 1,024. 
For the GARAN unit in RealGIN, the number of attentions $k$ is 4, and the attention dimension is 256. 
The filter size and channel number of the convolution layer after the GARAN unit and before the prediction layer are set to be $1\times 1$ and 1,024, respectively. 
The activation function for convolutional layers in the baseline model and RealGIN is LeakyReLU\cite{redmon2018yolov3:}, and the Batch Normalization is added after each convolutional layer. The hyper parameter $\lambda$ is 0.05.
During the training, the parameters of visual backbone are fixed. 
The batch size is 64, and the learning rate is set to be 0.001, which is halved for every 10 epochs. 
The max epoch is set to be 60, and the early stop is applied when the loss does not decay during 5 epochs.
The optimizer is Adam~\cite{kingma2014adam}.

\subsection{Experimental Results}

\subsubsection{Quantitative Analysis}
We first evaluate the effectiveness of each component in our model, of which results are shown in Tab.\ref{tab:ab_study}. 
The first block in Tab.\ref{tab:ab_study} presents different settings of our baseline model:
``\emph{visual only}'' denotes using the Yolo3~\cite{redmon2018yolov3:} as the REC model and predicting the target bounding box by selecting the one with the highest objectness score.
From these results, we observe that pretraining on object detection datasets is important to the model, which can provide good detection capabilities.
However, the result of ``\emph{visual only}'' also implies the importance of the language-vision modeling. 
The second block shows the efficacy of RealGIN's designs. 
As shown in Tab.\ref{tab:ab_study}, each design in RealGIN significantly improves the model performance.
In RefCOO+, where expressions are about object appearances, the improvement by AFS is up to 10.4\%, strongly suggesting that our design greatly enhances the visual features. 
In RefCOCOg with long expressions, GARAN with \emph{att\_loss} can also bring up to 4.1\% increase, which also confirms its advantage in the context reasoning. 
Meanwhile, the benefit of the attention loss is also significant.

In terms of the context modeling, we also compare the proposed GARAN unit to the alternatives.
Here, ``\emph{Baseline+Att}'' denotes using the attention feature to update the textual feature by following the setting in \cite{yang2016stacked,xu2015show}.
``\emph{+Loss}'' means the use of attention loss proposed in Sec.\ref{sec:garan}.
``+\emph{Transformer}'' denotes using the multi-attention layer~\cite{vaswani2017attention} to model the context information, and $k$ denotes the number of attention heads.
As shown in Tab.\ref{tab:context}, under the same settings, GARAN is better than the alternatives, and its advance becomes more significant when combined with the attention loss, which greatly confirms the merit of the proposed GARAN.

In terms of the model performance and processing speed, we compare the proposed baseline model and RealGIN with the SoA methods on RefCOCO, RefCOCO+ and RefCOCOg, the results of which are shown in Tab.\ref{tab:sota_results}. 
From Tab.\ref{tab:sota_results}, we can first observe that our baseline model is already better than most SoA methods, validating the feasibility of one-stage REC.
Its performance on the location-related dataset, \emph{i.e.,} RefCOCO, also shows its intrinsic advantages in spatial information modeling. 
To explain, prior multi-stage methods typically perform language-vision matching between feature vectors of the region and the expression, resulting in the lost of spatial information. 
In contrast, in one-stage inference, the model directly predicts the bounding box from the feature matrix, where the regional structure and spatial information are well preserved. 
Another important observation is that RealGIN has achieved very close performance to the most advanced SoA method, \emph{i.e.,} \emph{MAttNet}, on all three datasets. 
However, we also observe RealGIN's shortcoming in referring large instances like \emph{People} on Test A of RefCOCO+ compared to MAttNet. 
The main reason is that the region features used in MAttNet merit at describing more complete object information than that of CNN feature maps used in our scheme~\cite{anderson2018bottom}. 
Nevertheless, while retaining high performance, RealGIN significantly outperforms all SoA methods in terms of the computation efficiency. 
As shown in Tab.\ref{tab:sota_results}, the processing speed of SoA methods is only 1.3 to 2.1 FPS, while the one of RealGIN is 26.3 FPS, which is about 10 to 20 times faster than the existing methods.
In addition to the superior accuracy, this speed advantage also strongly confirms the contribution of this paper.
We further compare our method with SOTA on another two datasets, \emph{i.e.,} Flickr30k and ReferIT, as shown in Tab.\ref{tab:refit}.
The sizes of targets in these two datasets varies vastly, so we extend RealGIN to a multi-scale version by using the ong-stage model ZSGNet~\cite{sadhu2019zero-shot} as our base model. 
On these two datasets, our performance gains are more significant, greatly outperforming existing methods. 
Compared to ZSGNet, which also use the one-stage design, our merits validate the proposed designs again, \emph{i.e.,} AFS and GARAN.

\subsubsection{Qualitative Analysis}
We further visualize the predictions of the baseline and RealGIN to gain insights into both models, as shown in Fig.\ref{fig:figurefourvis}.
Fig.\ref{fig:figurefourvis}-(a) shows the comparisons between the baseline and RealGIN.
As shown in these examples, RealGIN can more confidently locate the target objects than the baseline.
Meanwhile, it also significantly merits in modeling expressions with attributes or multiple reference conditions, \emph{e.g.,} Exp.(2), (3), which subsequently validates the design of AFS and GARAN.
Fig.\ref{fig:figurefourvis}-(b) shows visualized attentions and grounding of RealGIN.
It can be seen that RealGIN is capable of referring short expressions with high confidence in images with complex visual scenes, \emph{e.g.,} Exp. (2)-(4). Such a result gives a high probability of practical application.
Meanwhile, it can also handle long expressions with complex linguistic conditions, showing a high reasoning ability in one-stage REC, \emph{e.g.,} Exp.(1), (4)-(6). 
Also, RealGIN is able to learn some interesting knowledge never appearing in pretraining, \emph{e.g.,} ``horse ass'' in Exp.(8), confirming its ability of modeling unseen categories. 
Besides, we have two findings from these results. 
First, the AFS scores also reflect the different properties of three datasets, \emph{e.g.,} the appearance-related RefCOCO+ has more requirements of low-level features. 
Second, the attention results also suggest that GARAN can capture all expression-related information towards the phase grounding.
These two findings further confirm the effectiveness of AFS and GARAN. 

Fig.\ref{fig:figurefourvis}-(c) gives some failure cases of RealGIN.
From these cases, we observe that in addition to the label noise and ambiguities, the main factor leading to failure prediction is the insufficient knowledge about visual concepts appearing in expressions.
Some visual concepts are very abstract and appear very few times in the dataset, causing the model to fail to master them well, such as \emph{sun shining} or \emph{the number 2}. 
Meanwhile, the backbone of RealGIN is pretrained with very few object categories, which also limits its recognition performance.
With more REC training examples and the pretraining with more categories, we believe that the performance of RealGIN can be further improved.  

\section{Conclusion}
In this paper, we address the computation inefficiency of Referring Expression Comprehension (REC).
All existing methods of REC follow a multi-stage pipeline, which is computation-ally expensive.
In contrast, we argue that real-time REC can be achieved by an one-stage method.
To validate this motivation, we first presented a simple baseline that embeds a language module to a one-stage object detection model, which still outperforms most prior works. 
To address two main issues of one-stage REC, \emph{i.e.,}, the diversity and complexity of expression, we further proposed the RealGIN network with two innovative designs, \emph{i.e.,} AFS and GARAN.
AFS adaptively fuses visual features at different semantic levels to represent the varying content in expressions, which greatly enhances the model discriminability.
GARAN uses the textual feature as a pivot to collect and diffuse query-aware information to all anchors, providing sufficient context for modeling linguistic conditions in expression.  
On three benchmark datasets, RealGIN achieves very competitive performance against the most advanced REC method: MAttNet. Most importantly, its processing speed is about 10 times faster than existing multi-stage methods.
{\small
	\bibliographystyle{ieee}
	\bibliography{egbib}

\begin{thebibliography}{10}\itemsep=-1pt

\bibitem{anderson2018bottom}
P.~Anderson, X.~He, C.~Buehler, D.~Teney, M.~Johnson, S.~Gould, and L.~Zhang.
\newblock Bottom-up and top-down attention for image captioning and visual
  question answering.
\newblock In {\em CVPR}, volume~3, page~6, 2018.

\bibitem{chen2017query-guided}
K.~Chen, R.~Kovvuri, and R.~Nevatia.
\newblock Query-guided regression network with context policy for phrase
  grounding.
\newblock {\em international conference on computer vision}, pages 824--832,
  2017.

\bibitem{dietterich1997solving}
T.~G. Dietterich, R.~H. Lathrop, and T.~Lozanoperez.
\newblock Solving the multiple instance problem with axis-parallel rectangles.
\newblock {\em Artificial Intelligence}, 89(1):31--71, 1997.

\bibitem{fukui2016multimodal}
A.~Fukui, D.~H. Park, D.~Yang, A.~Rohrbach, T.~Darrell, and M.~Rohrbach.
\newblock Multimodal compact bilinear pooling for visual question answering and
  visual grounding.
\newblock {\em arXiv preprint arXiv:1606.01847}, 2016.

\bibitem{hardoon2004canonical}
D.~R. Hardoon, S.~Szedmak, and J.~Shawetaylor.
\newblock Canonical correlation analysis: An overview with application to
  learning methods.
\newblock {\em Neural Computation}, 16(12):2639--2664, 2004.

\bibitem{hochreiter1997long}
S.~Hochreiter and J.~Schmidhuber.
\newblock Long short-term memory.
\newblock {\em Neural Computation}, 9(8):1735--1780, 1997.

\bibitem{hu2017modeling}
R.~Hu, M.~Rohrbach, J.~Andreas, T.~Darrell, and K.~Saenko.
\newblock Modeling relationships in referential expressions with compositional
  modular networks.
\newblock {\em computer vision and pattern recognition}, pages 4418--4427,
  2017.

\bibitem{hu2016natural}
R.~Hu, H.~Xu, M.~Rohrbach, J.~Feng, K.~Saenko, and T.~Darrell.
\newblock Natural language object retrieval.
\newblock {\em computer vision and pattern recognition}, pages 4555--4564,
  2016.

\bibitem{kazemzadeh2014referitgame:}
S.~Kazemzadeh, V.~Ordonez, M.~Matten, and T.~L. Berg.
\newblock Referitgame: Referring to objects in photographs of natural scenes.
\newblock pages 787--798, 2014.

\bibitem{kingma2014adam}
D.~Kingma and J.~Ba.
\newblock Adam: A method for stochastic optimization.
\newblock {\em arXiv preprint arXiv:1412.6980}, 2014.

\bibitem{krishna2018referring}
R.~Krishna, I.~Chami, M.~S. Bernstein, and L.~Feifei.
\newblock Referring relationships.
\newblock {\em computer vision and pattern recognition}, 2018.

\bibitem{krishna2016visual}
R.~Krishna, Y.~Zhu, O.~Groth, J.~Johnson, K.~Hata, J.~Kravitz, S.~Chen,
  Y.~Kalantidis, L.-J. Li, D.~A. Shamma, et~al.
\newblock Visual genome: Connecting language and vision using crowdsourced
  dense image annotations.
\newblock {\em arXiv preprint arXiv:1602.07332}, 2016.

\bibitem{li2017deep}
J.~Li, Y.~Wei, X.~Liang, F.~Zhao, J.~Li, T.~Xu, and J.~Feng.
\newblock Deep attribute-preserving metric learning for natural language object
  retrieval.
\newblock pages 181--189, 2017.

\bibitem{lin2017feature}
T.-Y. Lin, P.~Doll{\'a}r, R.~Girshick, K.~He, B.~Hariharan, and S.~Belongie.
\newblock Feature pyramid networks for object detection.
\newblock In {\em CVPR}, volume~1, page~4, 2017.

\bibitem{lin2014microsoft}
T.-Y. Lin, M.~Maire, S.~Belongie, J.~Hays, P.~Perona, D.~Ramanan,
  P.~Doll{\'a}r, and C.~L. Zitnick.
\newblock Microsoft coco: Common objects in context.
\newblock In {\em European conference on computer vision}, pages 740--755.
  Springer, 2014.

\bibitem{liu2017referring}
J.~Liu, L.~Wang, and M.~Yang.
\newblock Referring expression generation and comprehension via attributes.
\newblock {\em International Conference on Computer Vision}, pages 4866--4874,
  2017.

\bibitem{luo2017comprehension-guided}
R.~Luo and G.~Shakhnarovich.
\newblock Comprehension-guided referring expressions.
\newblock {\em computer vision and pattern recognition}, pages 3125--3134,
  2017.

\bibitem{mao2016generation}
J.~Mao, J.~Huang, A.~Toshev, O.~M. Camburu, A.~L. Yuille, and K.~P. Murphy.
\newblock Generation and comprehension of unambiguous object descriptions.
\newblock {\em computer vision and pattern recognition}, pages 11--20, 2016.

\bibitem{nagaraja2016modeling}
V.~K. Nagaraja, V.~I. Morariu, and L.~S. Davis.
\newblock Modeling context between objects for referring expression
  understanding.
\newblock {\em european conference on computer vision}, pages 792--807, 2016.

\bibitem{pennington2014glove}
J.~Pennington, R.~Socher, and C.~Manning.
\newblock Glove: Global vectors for word representation.
\newblock In {\em Proceedings of the 2014 conference on empirical methods in
  natural language processing (EMNLP)}, pages 1532--1543, 2014.

\bibitem{plummer2017conditional}
B.~A. Plummer, P.~Kordas, M.~H. Kiapour, S.~Zheng, R.~Piramuthu, and
  S.~Lazebnik.
\newblock Conditional image-text embedding networks.
\newblock {\em arXiv: Computer Vision and Pattern Recognition}, 2017.

\bibitem{plummer2017phrase}
B.~A. Plummer, A.~Mallya, C.~M. Cervantes, J.~Hockenmaier, and S.~Lazebnik.
\newblock Phrase localization and visual relationship detection with
  comprehensive image-language cues.
\newblock {\em international conference on computer vision}, pages 1946--1955,
  2017.

\bibitem{plummer2015flickr30k}
B.~A. Plummer, L.~Wang, C.~M. Cervantes, J.~C. Caicedo, J.~Hockenmaier, and
  S.~Lazebnik.
\newblock Flickr30k entities: Collecting region-to-phrase correspondences for
  richer image-to-sentence models.
\newblock {\em international conference on computer vision}, pages 2641--2649,
  2015.

\bibitem{redmon2018yolov3:}
J.~Redmon and A.~Farhadi.
\newblock Yolov3: An incremental improvement.
\newblock {\em arXiv: Computer Vision and Pattern Recognition}, 2018.

\bibitem{ren2017faster}
S.~Ren, K.~He, R.~Girshick, and J.~Sun.
\newblock Faster r-cnn: towards real-time object detection with region proposal
  networks.
\newblock {\em IEEE transactions on pattern analysis and machine intelligence},
  39(6):1137--1149, 2017.

\bibitem{rohrbach2016grounding}
A.~Rohrbach, M.~Rohrbach, R.~Hu, T.~Darrell, and B.~Schiele.
\newblock Grounding of textual phrases in images by reconstruction.
\newblock {\em european conference on computer vision}, pages 817--834, 2016.

\bibitem{sadhu2019zero-shot}
A.~Sadhu, K.~Chen, and R.~Nevatia.
\newblock Zero-shot grounding of objects from natural language queries.
\newblock {\em arXiv: Computer Vision and Pattern Recognition}, 2019.

\bibitem{teney2018tips}
D.~Teney, P.~Anderson, X.~He, and A.~V. Den~Hengel.
\newblock Tips and tricks for visual question answering: Learnings from the
  2017 challenge.
\newblock {\em computer vision and pattern recognition}, 2018.

\bibitem{vaswani2017attention}
A.~Vaswani, N.~Shazeer, N.~Parmar, J.~Uszkoreit, L.~Jones, A.~N. Gomez,
  L.~Kaiser, and I.~Polosukhin.
\newblock Attention is all you need.
\newblock {\em neural information processing systems}, pages 5998--6008, 2017.

\bibitem{wang2016learning}
L.~Wang, Y.~Li, and S.~Lazebnik.
\newblock Learning deep structure-preserving image-text embeddings.
\newblock {\em computer vision and pattern recognition}, pages 5005--5013,
  2016.

\bibitem{xu2015show}
K.~Xu, J.~Ba, R.~Kiros, K.~Cho, A.~C. Courville, R.~Salakhutdinov, R.~S. Zemel,
  and Y.~Bengio.
\newblock Show, attend and tell: Neural image caption generation with visual
  attention.
\newblock In {\em ICML}, volume~14, pages 77--81, 2015.

\bibitem{yang2016stacked}
Z.~Yang, X.~He, J.~Gao, L.~Deng, and A.~Smola.
\newblock Stacked attention networks for image question answering.
\newblock In {\em Proceedings of the IEEE Conference on Computer Vision and
  Pattern Recognition}, pages 21--29, 2016.

\bibitem{yu2018mattnet:}
L.~Yu, Z.~Lin, X.~Shen, J.~Yang, X.~Lu, M.~Bansal, and T.~L. Berg.
\newblock Mattnet: Modular attention network for referring expression
  comprehension.
\newblock {\em computer vision and pattern recognition}, pages 1307--1315,
  2018.

\bibitem{yu2016modeling}
L.~Yu, P.~Poirson, S.~Yang, A.~C. Berg, and T.~L. Berg.
\newblock Modeling context in referring expressions.
\newblock {\em european conference on computer vision}, pages 69--85, 2016.

\bibitem{yu2017a}
L.~Yu, H.~Tan, M.~Bansal, and T.~L. Berg.
\newblock A joint speaker-listener-reinforcer model for referring expressions.
\newblock {\em computer vision and pattern recognition}, pages 3521--3529,
  2017.

\bibitem{yu2018rethinking}
Z.~Yu, J.~Yu, C.~Xiang, Z.~Zhao, Q.~Tian, and D.~Tao.
\newblock Rethinking diversified and discriminative proposal generation for
  visual grounding.
\newblock {\em international joint conference on artificial intelligence},
  pages 1114--1120, 2018.

\bibitem{zhang2018grounding}
H.~Zhang, Y.~Niu, and S.~Chang.
\newblock Grounding referring expressions in images by variational context.
\newblock {\em computer vision and pattern recognition}, pages 4158--4166,
  2018.

\bibitem{zhang2017discriminative}
Y.~Zhang, L.~Yuan, Y.~Guo, Z.~He, I.~Huang, and H.~Lee.
\newblock Discriminative bimodal networks for visual localization and detection
  with natural language queries.
\newblock {\em computer vision and pattern recognition}, pages 1090--1099,
  2017.

\bibitem{zhuang2018parallel}
B.~Zhuang, Q.~Wu, C.~Shen, I.~D. Reid, and A.~V. Den~Hengel.
\newblock Parallel attention: A unified framework for visual object discovery
  through dialogs and queries.
\newblock {\em computer vision and pattern recognition}, pages 4252--4261,
  2018.

\end{thebibliography}
}

\newpage
\appendices
\section{Visualizations}
\begin{figure*}
	\centering
	\includegraphics[width=1\linewidth]{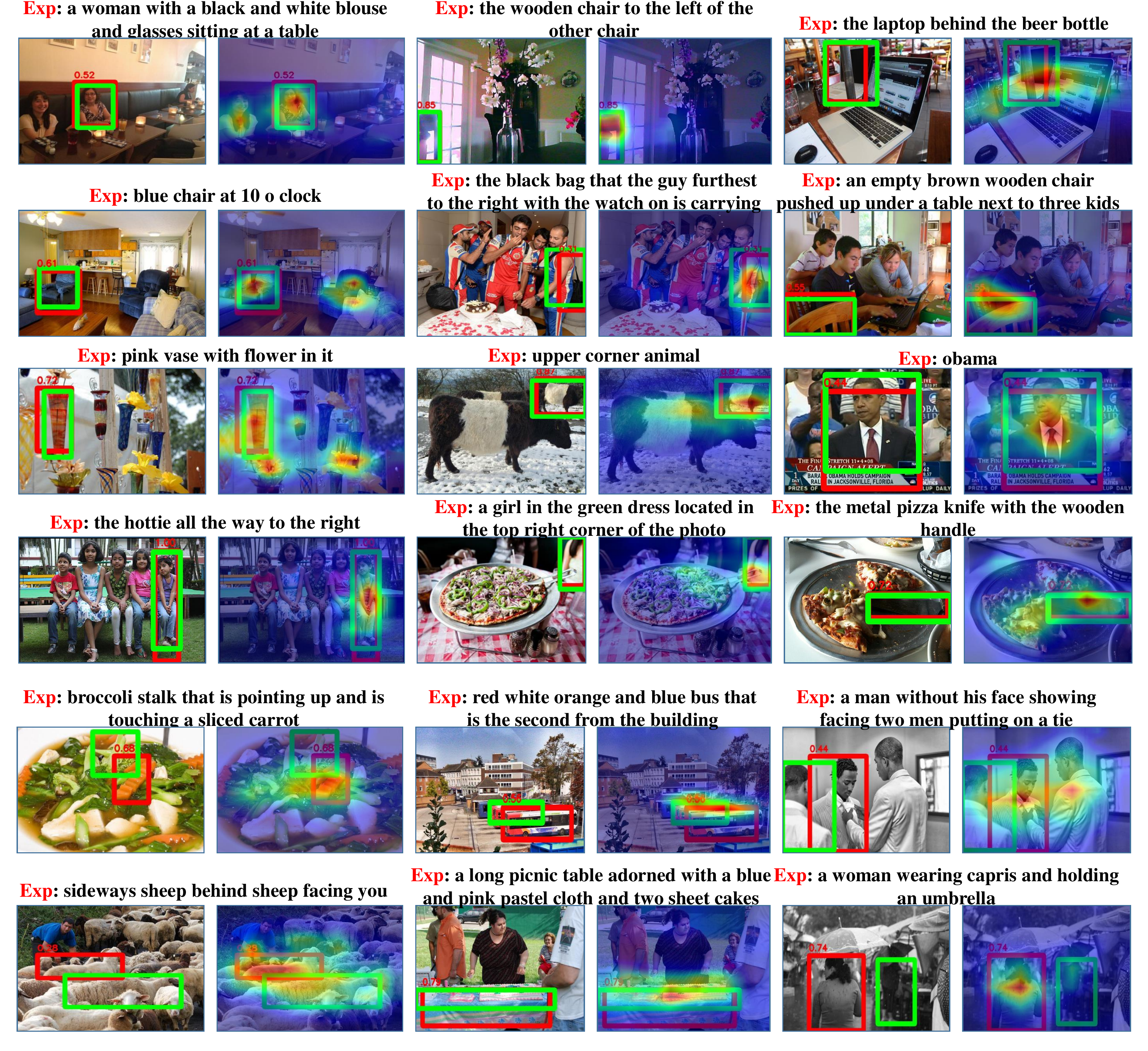}
	\caption{Predictions and attentions of RealGIN.
	}
	\label{fig:more_vis}
\end{figure*}




%




\end{document}